\documentclass{article}
\usepackage{hermitetwin_preprint,times}

\usepackage{amsmath,amsfonts,bm}

\def\eqref#1{equation~\ref{#1}}

\def\1{\bm{1}}

\DeclareMathAlphabet{\mathsfit}{\encodingdefault}{\sfdefault}{m}{sl}
\SetMathAlphabet{\mathsfit}{bold}{\encodingdefault}{\sfdefault}{bx}{n}

\newcommand{\R}{\mathbb{R}}

\usepackage{amsmath,amssymb,bm}
\usepackage{booktabs}
\usepackage{float}
\usepackage{graphicx}
\usepackage{xcolor}
\usepackage{hyperref}
\usepackage{url}
\hypersetup{
  colorlinks=true,linkcolor=blue,citecolor=blue,urlcolor=blue,
  pdftitle={HermiteTwin: Bicubic Hermite Shell Digital Twins for Contact-Rich Deformable Manipulation},
  pdfauthor={Kangrui Zhang, Mu Huang, Xing Shen, Tiantian Liu, Bo Ren}
}

\title{HermiteTwin: Bicubic Hermite Shell Digital Twins for Contact-Rich Deformable Manipulation}

\author{%
  \textbf{Kangrui Zhang}\textsuperscript{1}\quad
  \textbf{Mu Huang}\textsuperscript{2}\quad
  \textbf{Xing Shen}\textsuperscript{3}\quad
  \textbf{Tiantian Liu}\textsuperscript{4}\quad
  \textbf{Bo Ren}\textsuperscript{1}\\[0.35em]
  \textsuperscript{1}Nankai University\qquad
  \textsuperscript{2}Fudan University\\
  \textsuperscript{3}Shanghai Artificial Intelligence Laboratory\qquad
  \textsuperscript{4}Independent Researcher%
}

\newcommand{\method}{HermiteTwin}

\newcommand{\q}{\bm{q}}
\newcommand{\x}{\bm{x}}
\newcommand{\y}{\bm{y}}
\newcommand{\thetaf}{\bm{\theta}}

\begin{document}

\maketitle

\begin{abstract}
While existing methods for reconstructing deformable object manipulation have made impressive progress, they typically represent a thin object with a single particle set that must serve two purposes at once: modeling bending, and resolving contact with the robot. To bridge this gap, we present \method, a framework that reconstructs a physical digital twin of a thin deformable object from multiview video of a robot manipulating it. Our key idea is to represent deformation and contact separately, each at its own resolution. The object is simulated as a bicubic Hermite shell, a coarse grid of $C^1$ elements whose nodal positions and derivatives define a smooth surface with Kirchhoff--Love bending, while the gripper acts through dense samples in material space that are captured, clamped, and released, and whose attachment forces reach the shell through the Hermite basis. This decoupling keeps the physical state compact while contact stays fine-grained. We identify effective material and contact parameters with CMA-ES followed by adjoint-based refinement, and drive a reconstructed Gaussian appearance model with the simulated shell. In experiments on real cloth and foam manipulation, \method\ outperforms the state-of-the-art spring--mass twin across reconstruction, resimulation, and future prediction.
\end{abstract}

\section{Introduction}
\suppressfloats[t]

Robots that fold, lift, or spread cloth would benefit from a simulator that behaves like the specific object in front of them. Physical digital twins reconstructed from video provide such simulators: they recover the geometry, dynamics, and appearance of an object from a recorded interaction, and can then resimulate that interaction, predict its continuation, or render the outcome of new actions~\citep{jiang2025phystwin}. Simulations of this kind are already used to evaluate robot policies without physical trials~\citep{zhang2025real2sim}. How faithful a twin can be depends on two modeling choices: the physical representation of the object, and the way contact with the robot enters that representation.

\begin{figure}[t]
    \centering
    \includegraphics[width=\linewidth]{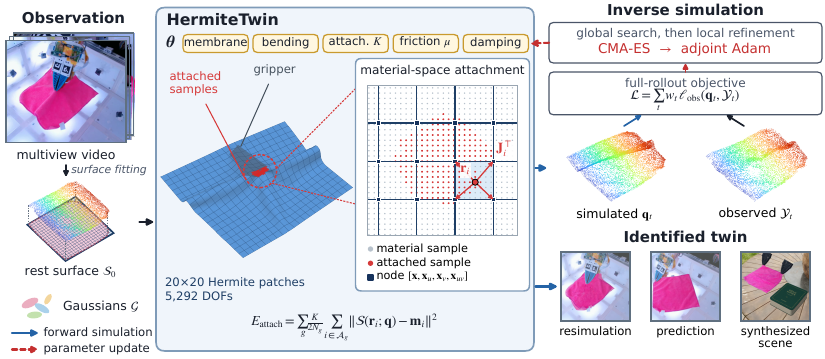}
    \caption{\textbf{Overview of \method.} From multiview video of a robot manipulating cloth, we fit a rest surface $\mathcal{S}_0$ to the first-frame object points and reconstruct a Gaussian appearance model $\mathcal{G}$. \method\ simulates the cloth as a coarse bicubic Hermite shell whose nodes store positions and derivatives. The gripper captures dense samples in material space (red); each attached sample $\bm{r}_i$ lies in one element, and its attachment residual reaches the 16 nodal vectors of that element through the Hermite weights $\bm{J}_i^{\top}$ (inset). The effective parameters $\thetaf$ are identified by minimizing a full-rollout observation loss with CMA-ES followed by adjoint-based Adam. The identified twin resimulates the recorded motion, predicts held-out frames, and drives the Gaussian appearance in synthesized scenes.}
    \label{fig:overview}
\end{figure}

Thin deformable objects make both choices difficult. Cloth resists stretching but bends easily, so its shape is dominated by curvature, and a gripper acts on it through a small patch of material that is captured, clamped, and later released. PhysTwin~\citep{jiang2025phystwin} handles both with one spring--mass system, whose particles represent the object and also connect it to the gripper. We present \method, a digital twin for thin deformable objects that represents deformation and contact separately, each at its own resolution (Figure~\ref{fig:overview}). Its physical state is a coarse bicubic Hermite shell~\citep{ni2024bhem}, a smooth ($C^1$) surface on which Kirchhoff--Love bending is defined directly and has its own stiffness. Contact, in contrast, is resolved in material space: the gripper captures, clamps, and releases dense material samples, each of which acts on the shell through the Hermite basis. The grasp can therefore be resolved finely while the shell itself stays coarse.

From multiview video of a robot interaction, we reconstruct the object's rest shape and a Gaussian appearance model~\citep{kerbl2023gaussians}, identify effective material and contact parameters by inverse simulation, and render the simulated shell by moving the Gaussians with it. Grasp events make the identification objective nonconvex and only piecewise smooth, so we combine global search with CMA-ES~\citep{hansen2001cmaes} and local refinement with adjoint gradients.

We evaluate \method\ on real robot manipulation from the Deform360 dataset~\citep{li2026deform360}, comparing it with PhysTwin and the learned point-based simulator GS-Dynamics~\citep{zhang2024gsdynamics}. \method\ reproduces the observed motion more accurately in 3D and in images and predicts the continuation of the recorded interactions more accurately. On a diagnostic interaction, transferring our grasp model to PhysTwin's particles does not recover the observed fold, which suggests that the shell representation, not only the grasp model, contributes to the difference. The identified twins can also be driven by new actions, such as draping the cloth over a book or placing it inside a box. In summary, our contributions are:
\begin{itemize}
    \item a physical digital twin for thin deformable objects built on a bicubic Hermite shell, a compact $C^1$ state with explicit Kirchhoff--Love bending;
    \item a material-space grasp model that captures, clamps, and releases dense material samples and couples them to the shell, so that the grasp is resolved finely on a coarse shell;
    \item an identification pipeline combining CMA-ES with adjoint refinement, and an evaluation on real manipulation data showing that \method\ resimulates observed motion and predicts future frames more accurately than spring--mass and learned baselines.
\end{itemize}

\section{Related Work}

\paragraph{Physical digital twins of deformable objects.}
PhysTwin~\citep{jiang2025phystwin} reconstructs a spring--mass model and Gaussian appearance from sparse-view videos and evaluates reconstruction, future prediction, and generalization to new interactions. Gaussian-splatting simulations of soft-body interactions have also been used to evaluate robot policies~\citep{zhang2025real2sim}. Learning-based simulators predict the motion of particles or Gaussians directly: GS-Dynamics~\citep{zhang2024gsdynamics} and ParticleFormer~\citep{huang2025particleformer} learn point-based dynamics, and SoMA~\citep{huang2026soma} learns an action-conditioned Gaussian simulator for robotic manipulation. PhysReal~\citep{deng2026physreal} combines analytical and neural constitutive models. PhysHanDI~\citep{lee2026physhandi} drives a spring--mass object with dense hand vertices through virtual springs and assumes that the hand--object contact topology stays fixed within a sequence; our grasp samples are instead captured and released as the gripper closes and opens. These methods address a broad range of objects. We focus on thin sheets that can be parameterized over a single rectangle, for which a shell state and material-space contact are natural.

\paragraph{Gaussian appearance for deformable objects.}
3D Gaussian Splatting~\citep{kerbl2023gaussians} represents scenes with anisotropic primitives that can be rendered efficiently and differentiably. PhysAvatar~\citep{zheng2024physavatar} and Gaussian Garments~\citep{rong2024gaussiangarments} couple simulated garment geometry with appearance reconstructed from multiview video, and GeoSplatting~\citep{ye2025geosplatting} uses explicit geometry to guide physically based inverse rendering. Cloth-Splatting~\citep{longhini2024clothsplatting} estimates cloth states from RGB supervision, and GauSim~\citep{shao2024gausim} simulates elastic objects with continuum-inspired Gaussian kernels. We keep appearance separate from mechanics: the Gaussians are transported by the simulated shell, and no image loss enters parameter identification, so the same appearance model can be driven by different simulators and by new actions.

\paragraph{Thin shells and differentiable simulation.}
Bicubic Hermite elements~\citep{ni2024bhem} provide $C^1$ continuity through explicit derivative degrees of freedom and model Kirchhoff--Love thin shells on coarse quadrilateral grids. B-spline shells also provide $C^1$ continuity and embed a contact mesh whose resolution can be chosen independently of the shell~\citep{meng2026bspline}. We use a Hermite shell as the state of a digital twin and apply separate material-space sampling to robot grasps, including capture and release. Collisions are handled with incremental potential contact~\citep{li2020ipc}. Differentiable simulators such as DiffTaichi~\citep{hu2020difftaichi} enable gradient-based identification, but grasp events make our objective only piecewise smooth. We therefore combine derivative-free search~\citep{hansen2001cmaes} with adjoint gradients computed for a fixed contact configuration.

\section{HermiteTwin}
\label{sec:method}

\subsection{Problem Setup}

We are given calibrated multiview videos of a robot manipulating an object and the gripper trajectory $\bm{a}_{0:T}$ with its clamp and release schedule. From these inputs we build a twin
\begin{equation}
    \mathcal{T}=\left(\mathcal{S}_0,\thetaf,\mathcal{A},\mathcal{G}\right),
\end{equation}
which consists of a rest surface $\mathcal{S}_0$ in material coordinates, effective material and contact parameters $\thetaf$, a grasp model $\mathcal{A}$ that maps gripper motion to forces on the material, and a Gaussian appearance model $\mathcal{G}$ (Figure~\ref{fig:overview}). We reconstruct $\mathcal{S}_0$ and $\mathcal{G}$ from the first frame and identify $\thetaf$ by inverse simulation. Given an action sequence, the twin produces a physical trajectory $\q_{0:T}$ and rendered images. As in PhysTwin, we identify one twin per recorded interaction. The parameters are inferred from video without force measurements, so we treat them as effective values: stiffness, rest shape, friction, and attachment can partially compensate for one another.

\subsection{Bicubic Hermite Shell}

We parameterize the object by material coordinates $(u,v)\in[0,1]^2$ on a structured grid of bicubic Hermite elements. The coordinates come from the minimum-area rectangle around the first-frame object points projected onto their fitted plane, and the sheet rests flat on this rectangle. Each node stores a position together with its first and mixed derivatives,
\begin{equation}
    \q_{ij}=\left[\x_{ij},\;\x_{u,ij},\;\x_{v,ij},\;\x_{uv,ij}\right]\in\R^{12}.
\end{equation}
Within an element, the surface is
\begin{equation}
    \mathcal{S}(u,v;\q)=\sum_{a=0}^{1}\sum_{b=0}^{1}
    \bm{H}_{a}(u)^{\top}\,\bm{Q}_{ab}\,\bm{H}_{b}(v),
    \label{eq:hermite}
\end{equation}
where $\bm{H}$ denotes the cubic Hermite basis and $\bm{Q}_{ab}$ collects the position, tangent, and twist vectors of the element's corner nodes. Neighboring elements share nodal derivatives, so the surface is $C^1$ across element boundaries. The surface has continuous tangents and normals, and its curvature within each element defines the bending energy. The elastic energy combines a membrane term and a Kirchhoff--Love bending term,
\begin{equation}
    E_{\mathrm{el}}(\q;\thetaf)=s_{\mathrm{mem}}\,E_{\mathrm{mem}}(\q;E,\nu,h)+s_{\mathrm{bend}}\,E_{\mathrm{bend}}(\q;E,\nu,h),
    \label{eq:elastic}
\end{equation}
with nominal Young's modulus $E$, Poisson ratio $\nu$, thickness $h$, and a St.~Venant--Kirchhoff constitutive law. The scales $s_{\mathrm{mem}}$ and $s_{\mathrm{bend}}$ are part of $\thetaf$ and adjust the membrane and bending stiffness.

\paragraph{Time integration.}
We advance the shell implicitly by minimizing an incremental potential at each step,
\begin{equation}
    \q_{t+1}=\arg\min_{\q}\;
    \frac{1}{2\Delta t^{2}}\left\|\q-\hat{\q}_{t+1}\right\|_{\bm{M}}^{2}
    +E_{\mathrm{el}}(\q;\thetaf)
    +E_{\mathrm{attach}}(\q;\mathcal{M}_t)
    +E_{\mathrm{IPC}}(\q)
    +E_{\mathrm{fric}}(\q),
    \label{eq:incremental}
\end{equation}
where $\hat{\q}_{t+1}$ is the inertial prediction, $\bm{M}$ is the mass matrix, $E_{\mathrm{IPC}}$ is the barrier of incremental potential contact~\citep{li2020ipc} that prevents interpenetration with the gripper and the table, and $E_{\mathrm{fric}}$ is a lagged friction potential; we omit the damping term for brevity. Each step is solved with Newton's method and a line search. The attachment energy $E_{\mathrm{attach}}$ couples the shell to the gripper and is defined next.

\subsection{Material-Space Grasp Coupling}
\label{sec:grasp}

\paragraph{Dense material samples.}
We model the grasp with dense samples on a fixed grid in material space. Sample $i$ has a persistent material coordinate $\bm{r}_i=(u_i,v_i)$ and the physical position
\begin{equation}
    \x_i(\q)=\mathcal{S}(\bm{r}_i;\q).
    \label{eq:contact_sample}
\end{equation}
Samples are tied to material coordinates rather than snapped to Hermite nodes, and they are far denser than the nodes. Each sample is interpolated from the 16 nodal vectors (positions, tangents, and twists) of its element, so contact can be resolved densely around a fold without adding a particle for every sample. Contact resolution is thereby decoupled from state resolution: where a grasp-induced fold spans only a few elements, the attachment targets still cover it densely.

\paragraph{Capture, clamp, and release.}
Each sample is free, captured, or clamped. A free sample is captured when a thin box around a gripper pad sweeps over it during a substep. The capture records the sample's material coordinate, the gripper side that captured it, and an attachment target $\bm{m}_i$ expressed in the gripper frame. Before clamp, the target may relax tangentially toward the sample; at clamp, it is frozen in the gripper frame and moves rigidly with the gripper. Release, or moving out of the pad region before clamp, returns the sample to the free state, from which it can be captured again. Because the grasp is defined on material identities, a new grasp acquires whatever material lies under the gripper at that time rather than a fixed set of nodes.

\paragraph{Attachment energy.}
For gripper $g$ with a nonempty set $\mathcal{A}_g$ of active samples and $N_g=|\mathcal{A}_g|$, the attachment energy is a normalized isotropic penalty,
\begin{equation}
    E_{\mathrm{attach}}(\q;\mathcal{M})=
    \sum_g\frac{K}{2N_g}\sum_{i\in\mathcal{A}_g}
    \left\|\mathcal{S}(\bm{r}_i;\q)-\bm{m}_i\right\|^2,
    \label{eq:sample_attach_energy}
\end{equation}
where $\mathcal{M}$ denotes the active sets and targets and $K$ is the attachment stiffness. Normalizing by $N_g$ keeps the total stiffness of each grasp independent of the sample density. Because $\mathcal{S}$ is linear in $\q$, the Jacobian $\bm{J}_i=\partial\mathcal{S}(\bm{r}_i;\q)/\partial\q$ consists of the Hermite basis weights of sample $i$, and the residual of each sample acts on the 16 nodal vectors of its element. For a fixed active set, $E_{\mathrm{attach}}$ is quadratic in $\q$ and adds a constant positive semidefinite term to the Newton system of Eq.~(\ref{eq:incremental}). Dense contact coverage therefore adds no degrees of freedom: the physical state remains a compact Hermite surface.

\subsection{Inverse Simulation}

We identify the parameters by minimizing an observation loss over a full rollout,
\begin{equation}
    \min_{\thetaf}\;\mathcal{L}(\thetaf)=
    \sum_{t\in\mathcal{O}}w_t\,\ell_{\mathrm{obs}}(\q_t,\mathcal{Y}_t)
    \quad \text{s.t.}\quad \q_{t+1}=F(\q_t,\dot{\q}_t;\thetaf,\bm{a}_t),
    \label{eq:inverse}
\end{equation}
where $F$ is one step of Eq.~(\ref{eq:incremental}), $\mathcal{O}$ is the set of fitting frames with equal weights $w_t=1/|\mathcal{O}|$, and $\mathcal{Y}_t$ is the set of tracked object points at frame $t$. The observation term is the squared bidirectional Chamfer distance
\begin{equation}
    \ell_{\mathrm{obs}}(\q_t,\mathcal{Y}_t)=
    \frac{1}{|\mathcal{Y}_t|}\sum_{\y\in\mathcal{Y}_t}\min_{\x\in\mathcal{X}_t}\|\y-\x\|_2^2
    +\frac{1}{|\mathcal{X}_t|}\sum_{\x\in\mathcal{X}_t}\min_{\y\in\mathcal{Y}_t}\|\x-\y\|_2^2,
    \label{eq:geomloss}
\end{equation}
where $\mathcal{X}_t=\mathcal{X}(\q_t)$ are simulated points at fixed material locations (Section~\ref{sec:setup}). Rather than $E$, $\nu$, and $h$ individually, we optimize a few effective parameters in log space: the stiffness scales $s_{\mathrm{mem}}$ and $s_{\mathrm{bend}}$ of Eq.~(\ref{eq:elastic}), the attachment stiffness $K$, table friction, and damping.

Capture and clamp events make $\mathcal{L}$ nonconvex and only piecewise smooth. We therefore first search the parameters with CMA-ES~\citep{hansen2001cmaes}, replaying the same clamp and release schedule for every candidate, and then refine the best candidate with Adam using discrete-adjoint gradients. At each step, the converged Newton residual $\bm{f}_{t+1}(\q_{t+1},\q_t;\thetaf)=\bm{0}$ defines $\q_{t+1}$ implicitly; the adjoint solves linear systems with $\partial\bm{f}_{t+1}/\partial\q_{t+1}$ backward in time and accumulates $\mathrm{d}\mathcal{L}/\mathrm{d}\thetaf$ with the contact configuration held fixed. Section~\ref{sec:analysis} compares such gradients with finite differences when the grasp relies on frictional contact instead.

\subsection{Appearance Transport and New Actions}

We reconstruct Gaussians~\citep{kerbl2023gaussians} of the object in the first frame and bind each to its 16 nearest surface points in that frame (material samples for \method, particles for PhysTwin). Its mean and covariance follow the local rigid motion of these neighbors, estimated by Procrustes fitting as in the linear blend skinning of PhysTwin~\citep{jiang2025phystwin}. Colors, opacities, and reference anisotropies stay fixed, and no image loss is used to identify $\thetaf$. The same appearance model can therefore be driven by any simulated trajectory: by either simulator in our comparison, or by new actions. For a new action, we replace the recorded gripper trajectory and continue the simulation from the identified state, which includes positions and velocities, without re-initializing the object. Embedding each Gaussian directly in the continuous Hermite surface is a natural alternative; we keep the neighbor-based rule so that both simulators share one appearance model.

\section{Experiments}
\label{sec:experiments}

We evaluate how accurately \method\ resimulates observed manipulation (Section~\ref{sec:recon}), how well it predicts held-out future frames (Section~\ref{sec:future}), which components contribute to its accuracy (Section~\ref{sec:analysis}), and how it supports new interactions (Section~\ref{sec:applications}).

\subsection{Experimental Setup}
\label{sec:setup}

\paragraph{Data.}
We use ten interactions from Deform360~\citep{li2026deform360}, a multiview visuotactile dataset of robot manipulation: eight with a red cloth and two with a white foam sheet (Appendix~\ref{app:dataset}). For each interaction, object points with fixed identities tracked over time serve as the observations $\mathcal{Y}_t$, together with the recorded robot motion and tactile signals. Contact onset, clamp, and release are extracted from the tactile signals and frozen before fitting; PhysTwin uses the same onset and release.

\paragraph{Baselines.}
We compare with PhysTwin~\citep{jiang2025phystwin}, keeping its spring--mass model, integrator, and optimizer. Because the gripper approaches the object from a distance in these recordings, its controller springs are connected once, to the object points near the gripper at contact onset, and act only between onset and release. We also compare with GS-Dynamics~\citep{zhang2024gsdynamics}, a learned point-based simulator that predicts the motion of the object points from the gripper motion with a graph neural network. Within each interaction, all methods use the same observations and transport the same Gaussian appearance model, so differences in the rendered images reflect differences in the simulated motion.

\paragraph{Fitting.}
\method\ and PhysTwin fit every interaction with the objective of Eq.~(\ref{eq:geomloss}), which also selects the model; each starts from its default parameters and runs CMA-ES followed by Adam refinement (Appendix~\ref{app:implementation}), and for PhysTwin, this objective replaces its native loss. GS-Dynamics is trained on the same fitting frames with its native loss, and we keep the checkpoint whose rollout over these frames has the lowest Chamfer distance. All methods are scored with Eq.~(\ref{eq:geomloss}) at a common set of material locations obtained from PhysTwin's particle sampling of the first frame: \method\ predicts the object at these locations, PhysTwin's particles contain them, and GS-Dynamics predicts all tracked points, which include them.

\paragraph{Metrics.}
We report the objective of Eq.~(\ref{eq:geomloss}) as the 3D error (CD, in mm$^2$) and measure image accuracy with silhouette IoU and PSNR in three fixed camera views; for prediction, we also report SSIM and LPIPS. Renderings are compared with the recorded images on a black background, excluding gripper pixels. Metrics are averaged with equal weights over views, then frames, then interactions.

\subsection{Reconstruction \& Resimulation}
\label{sec:recon}

\begin{figure}[t]
    \centering
    \setlength{\fboxsep}{0pt}
    \setlength{\fboxrule}{0.4pt}
    \setlength{\tabcolsep}{0.8pt}
    \renewcommand{\arraystretch}{0.6}
    \newcommand{\rowlabel}[1]{\rotatebox{90}{\makebox[1.86cm]{\scriptsize #1}}}
    \begin{tabular}{@{}c@{\hspace{2pt}}ccc@{\hspace{6pt}}ccc@{}}
        & \multicolumn{3}{c}{\small (a) Red cloth} & \multicolumn{3}{c}{\small (b) White foam sheet} \\[2pt]
        & \scriptsize $t=100$ & \scriptsize $t=160$ & \scriptsize $t=269$ & \scriptsize $t=100$ & \scriptsize $t=150$ & \scriptsize $t=240$ \\
        \rowlabel{Observation} &
        \fcolorbox{blue!55!black}{white}{\includegraphics[height=1.86cm]{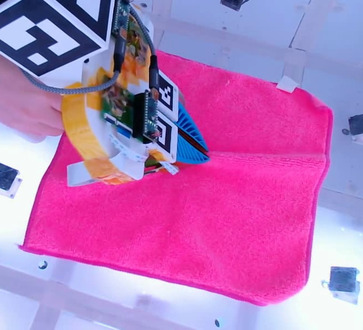}} &
        \fcolorbox{blue!55!black}{white}{\includegraphics[height=1.86cm]{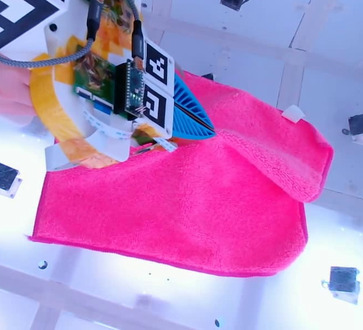}} &
        \fcolorbox{blue!55!black}{white}{\includegraphics[height=1.86cm]{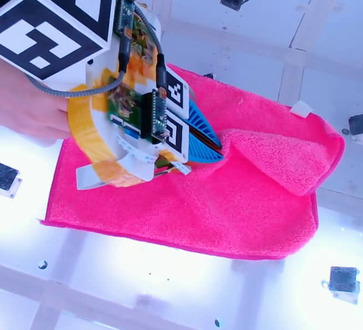}} &
        \fcolorbox{blue!55!black}{white}{\includegraphics[height=1.86cm]{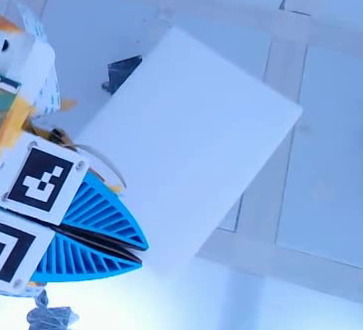}} &
        \fcolorbox{blue!55!black}{white}{\includegraphics[height=1.86cm]{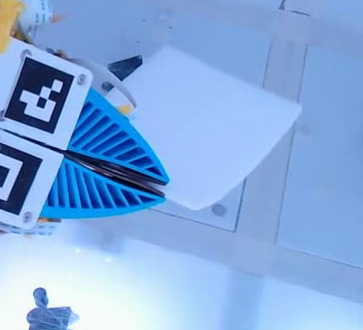}} &
        \fcolorbox{blue!55!black}{white}{\includegraphics[height=1.86cm]{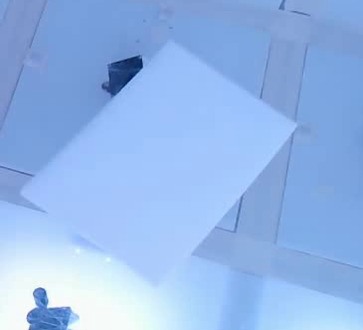}} \\
        \rowlabel{\method} &
        \fcolorbox{blue!55!black}{white}{\includegraphics[height=1.86cm]{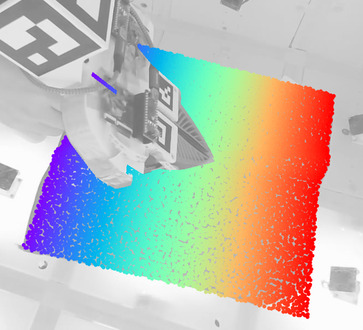}} &
        \fcolorbox{blue!55!black}{white}{\includegraphics[height=1.86cm]{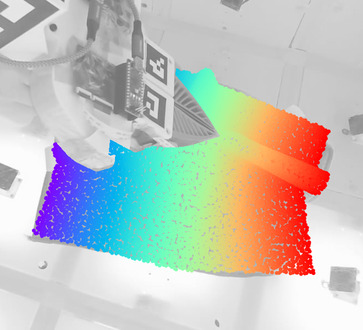}} &
        \fcolorbox{blue!55!black}{white}{\includegraphics[height=1.86cm]{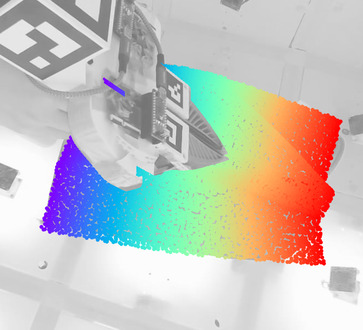}} &
        \fcolorbox{blue!55!black}{white}{\includegraphics[height=1.86cm]{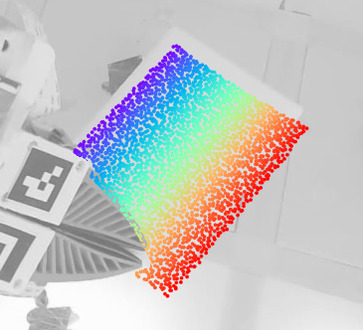}} &
        \fcolorbox{blue!55!black}{white}{\includegraphics[height=1.86cm]{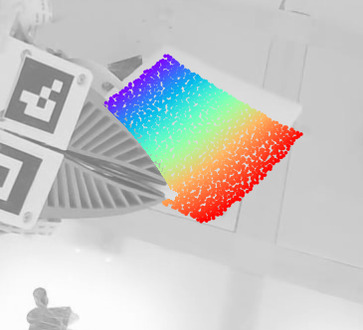}} &
        \fcolorbox{blue!55!black}{white}{\includegraphics[height=1.86cm]{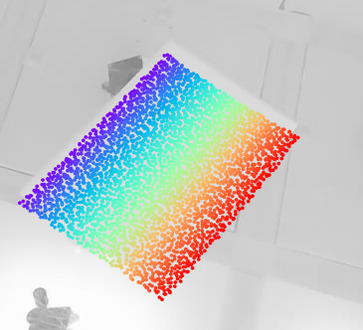}} \\
        \rowlabel{PhysTwin} &
        \fcolorbox{blue!55!black}{white}{\includegraphics[height=1.86cm]{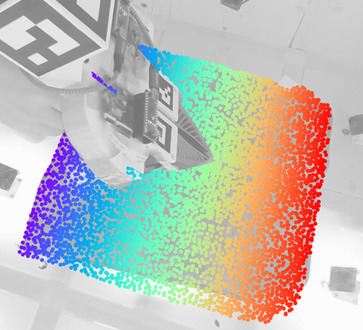}} &
        \fcolorbox{blue!55!black}{white}{\includegraphics[height=1.86cm]{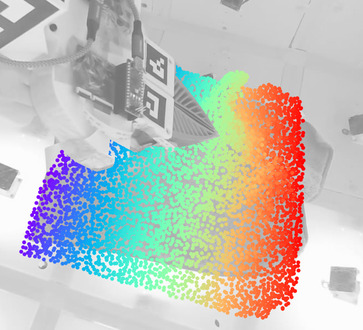}} &
        \fcolorbox{blue!55!black}{white}{\includegraphics[height=1.86cm]{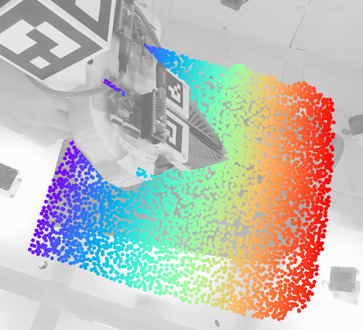}} &
        \fcolorbox{blue!55!black}{white}{\includegraphics[height=1.86cm]{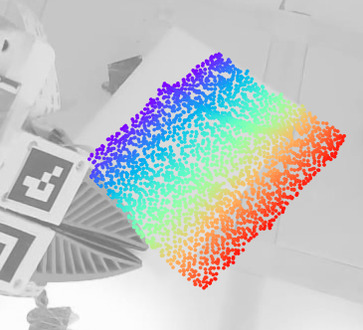}} &
        \fcolorbox{blue!55!black}{white}{\includegraphics[height=1.86cm]{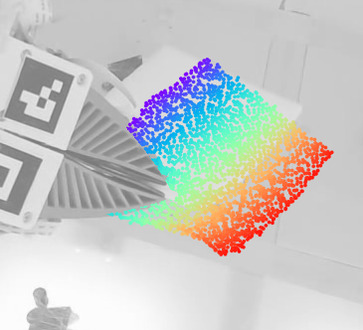}} &
        \fcolorbox{blue!55!black}{white}{\includegraphics[height=1.86cm]{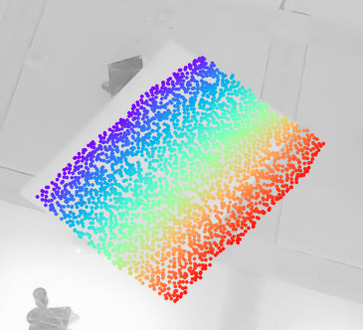}} \\
        \rowlabel{GS-Dynamics} &
        \fcolorbox{blue!55!black}{white}{\includegraphics[height=1.86cm]{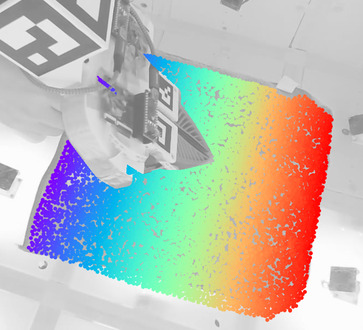}} &
        \fcolorbox{blue!55!black}{white}{\includegraphics[height=1.86cm]{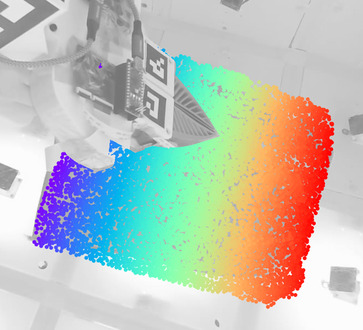}} &
        \fcolorbox{blue!55!black}{white}{\includegraphics[height=1.86cm]{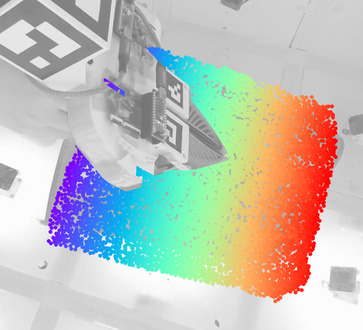}} &
        \fcolorbox{blue!55!black}{white}{\includegraphics[height=1.86cm]{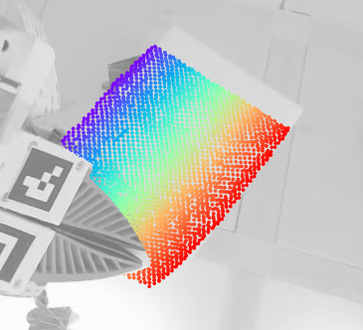}} &
        \fcolorbox{blue!55!black}{white}{\includegraphics[height=1.86cm]{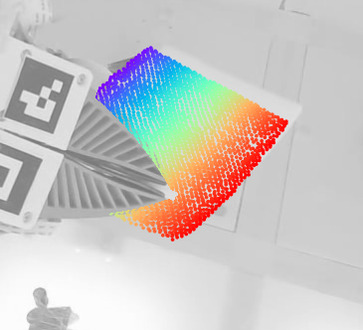}} &
        \fcolorbox{blue!55!black}{white}{\includegraphics[height=1.86cm]{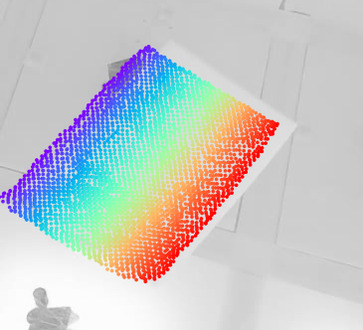}}
    \end{tabular}
    \caption{\textbf{Qualitative reconstruction and resimulation.} Observations and resimulations of the red cloth (a) and the foam sheet (b). Simulated panels show points sampled densely on the \method\ surface and the simulated PhysTwin and GS-Dynamics particles, colored by their material coordinate at rest. \method\ reproduces the folded cloth corner at $t=160$ and $t=269$.}
    \label{fig:cloth-fold-replay}
\end{figure}

Table~\ref{tab:reconstruction-resimulation} evaluates how well each twin resimulates the fitting frames of all ten interactions. \method\ improves all three metrics on average. Figure~\ref{fig:cloth-fold-replay} shows representative frames. The red-cloth interaction is dominated by contact and bending: the gripper pinches a corner, lifts it, and folds it back over the sheet. \method\ reproduces this fold at $t=160$ and $t=269$, whereas both baselines leave the sheet nearly flat. This holds even though PhysTwin is given the grasp timing from the tactile signals, so that its controller springs connect only once the gripper touches the cloth (Section~\ref{sec:setup}); the spring--mass sheet still does not form the fold. The foam sheet is much stiffer, and PhysTwin does not maintain this stiffness: its sheet deforms around the grasp instead of keeping its flat shape, whereas \method\ preserves it.

\begin{table}[!htb]
    \centering
    \caption{\textbf{Reconstruction and resimulation.} CD, three-view silhouette IoU, and PSNR over the fitting frames, averaged with equal weight over the ten interactions.}
    \label{tab:reconstruction-resimulation}
    \small
    \begin{tabular}{@{}lccc@{}}
        \toprule
        Method & CD $\downarrow$ (mm$^2$) & IoU $\uparrow$ (\%) & PSNR $\uparrow$ (dB) \\
        \midrule
        GS-Dynamics & 116.95 & 83.24 & 22.160 \\
        PhysTwin & 61.89 & 85.22 & 22.745 \\
        \method\ (ours) & \textbf{42.05} & \textbf{88.89} & \textbf{24.242} \\
        \bottomrule
    \end{tabular}
\end{table}

\subsection{Future Prediction}
\label{sec:future}

On the red cloth, the models fitted on the first 300 frames in Section~\ref{sec:recon} continue the same rollout to the end of each recording with frozen parameters; the simulated state is carried over without resetting it to the observation, and the held-out frames never enter fitting or model selection. The robot motion and contact events of the held-out frames come from the recording, so this experiment predicts the cloth under known future control rather than the robot's actions.

\begin{figure}[t]
    \centering
    \setlength{\fboxsep}{0pt}
    \setlength{\fboxrule}{0.4pt}
    \setlength{\tabcolsep}{0.8pt}
    \renewcommand{\arraystretch}{0.6}
    \newcommand{\rowlabel}[1]{\rotatebox{90}{\makebox[1.86cm]{\scriptsize #1}}}
    \begin{tabular}{@{}c@{\hspace{2pt}}ccc@{\hspace{6pt}}ccc@{}}
        & \multicolumn{3}{c}{\small Reconstruction \& Resimulation} & \multicolumn{3}{c}{\small Future Prediction} \\[2pt]
        & \scriptsize $t=100$ & \scriptsize $t=200$ & \scriptsize $t=300$ & \scriptsize $t=350$ & \scriptsize $t=410$ & \scriptsize $t=470$ \\
        \rowlabel{Observation} & \fcolorbox{blue!55!black}{white}{\includegraphics[height=1.86cm]{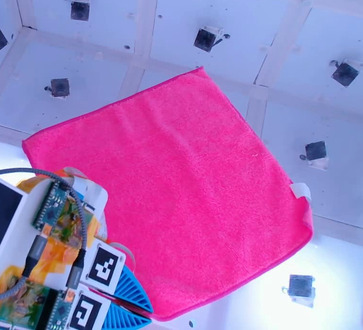}} &
        \fcolorbox{blue!55!black}{white}{\includegraphics[height=1.86cm]{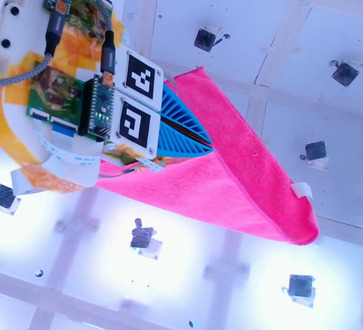}} &
        \fcolorbox{blue!55!black}{white}{\includegraphics[height=1.86cm]{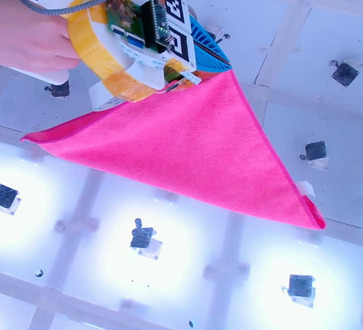}} &
        \fcolorbox{red!70!black}{white}{\includegraphics[height=1.86cm]{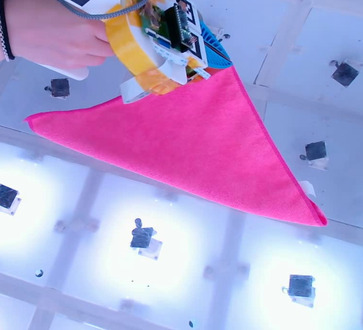}} &
        \fcolorbox{red!70!black}{white}{\includegraphics[height=1.86cm]{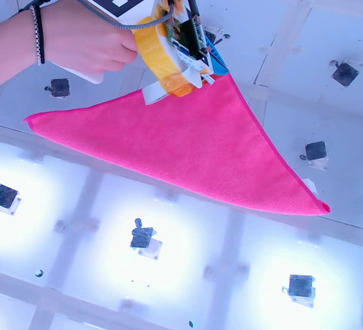}} &
        \fcolorbox{red!70!black}{white}{\includegraphics[height=1.86cm]{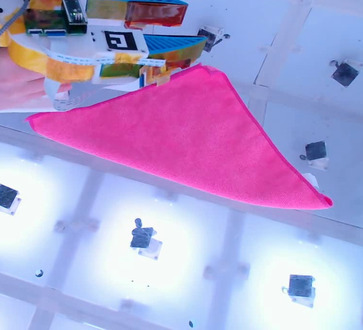}} \\
        \rowlabel{\method} & \fcolorbox{blue!55!black}{white}{\includegraphics[height=1.86cm]{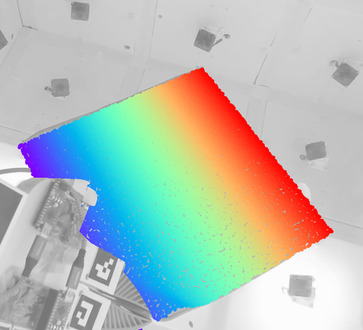}} &
        \fcolorbox{blue!55!black}{white}{\includegraphics[height=1.86cm]{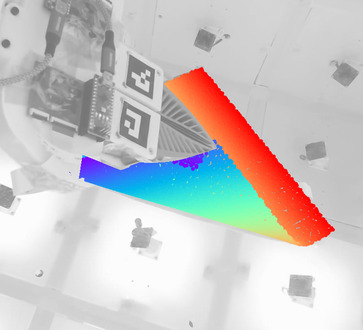}} &
        \fcolorbox{blue!55!black}{white}{\includegraphics[height=1.86cm]{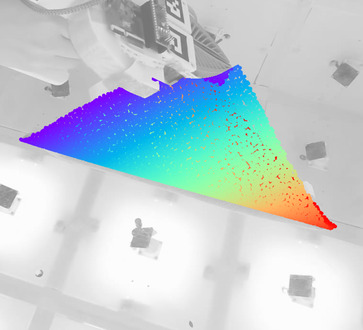}} &
        \fcolorbox{red!70!black}{white}{\includegraphics[height=1.86cm]{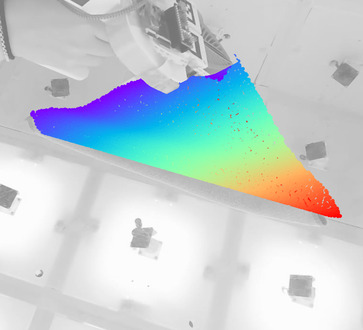}} &
        \fcolorbox{red!70!black}{white}{\includegraphics[height=1.86cm]{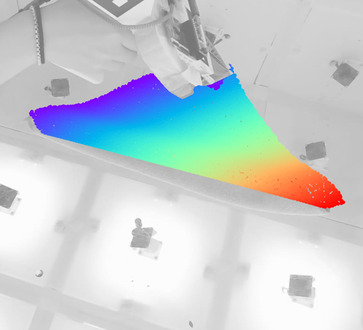}} &
        \fcolorbox{red!70!black}{white}{\includegraphics[height=1.86cm]{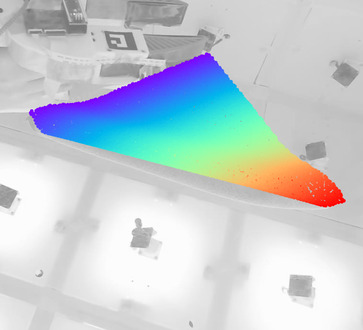}} \\
        \rowlabel{PhysTwin} & \fcolorbox{blue!55!black}{white}{\includegraphics[height=1.86cm]{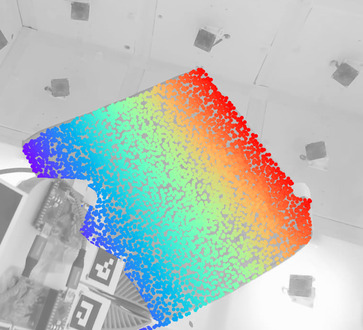}} &
        \fcolorbox{blue!55!black}{white}{\includegraphics[height=1.86cm]{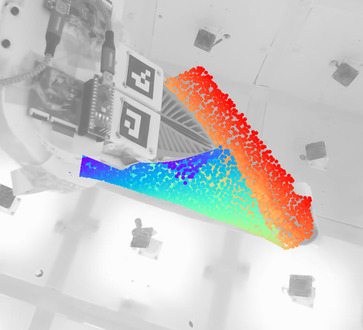}} &
        \fcolorbox{blue!55!black}{white}{\includegraphics[height=1.86cm]{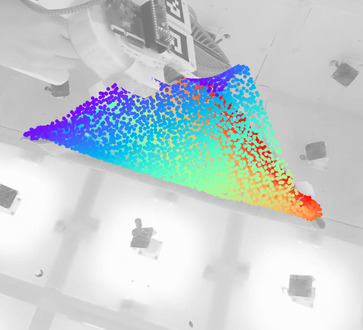}} &
        \fcolorbox{red!70!black}{white}{\includegraphics[height=1.86cm]{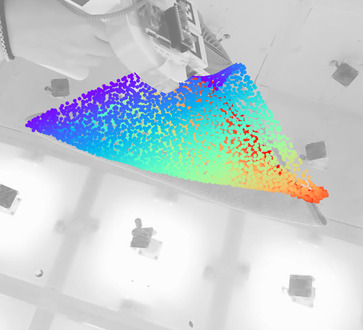}} &
        \fcolorbox{red!70!black}{white}{\includegraphics[height=1.86cm]{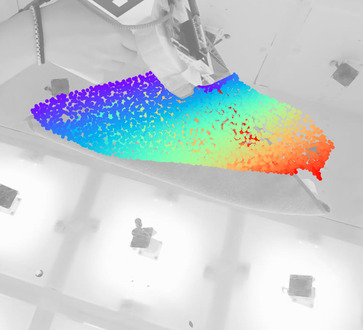}} &
        \fcolorbox{red!70!black}{white}{\includegraphics[height=1.86cm]{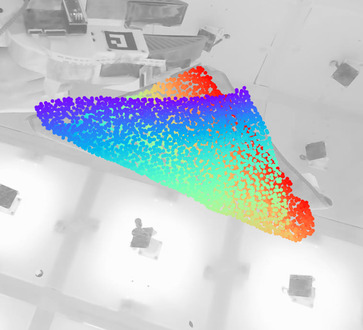}} \\
        \rowlabel{GS-Dynamics} & \fcolorbox{blue!55!black}{white}{\includegraphics[height=1.86cm]{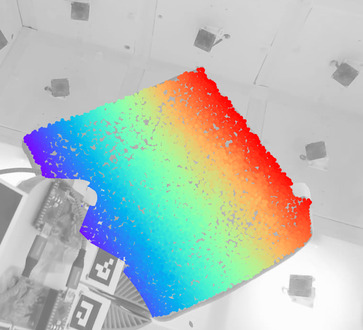}} &
        \fcolorbox{blue!55!black}{white}{\includegraphics[height=1.86cm]{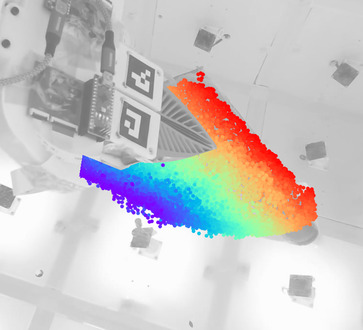}} &
        \fcolorbox{blue!55!black}{white}{\includegraphics[height=1.86cm]{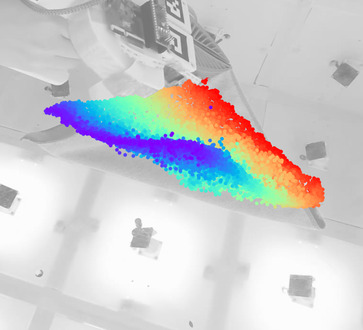}} &
        \fcolorbox{red!70!black}{white}{\includegraphics[height=1.86cm]{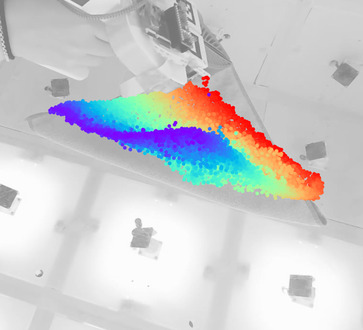}} &
        \fcolorbox{red!70!black}{white}{\includegraphics[height=1.86cm]{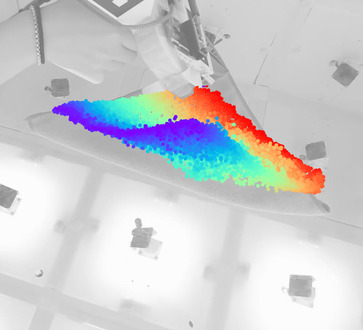}} &
        \fcolorbox{red!70!black}{white}{\includegraphics[height=1.86cm]{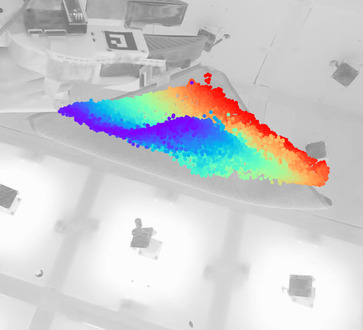}}
    \end{tabular}
    \caption{\textbf{Reconstruction and future prediction on a red-cloth interaction.} The first three columns show reconstruction and resimulation ($t=100,200,300$); the last three show future prediction ($t=350,410,470$), with parameters fitted on frames 1--300.}
    \label{fig:cloth-future-prediction}
\end{figure}

Table~\ref{tab:future-prediction} shows that \method\ predicts the held-out frames more accurately in all five metrics, and Figure~\ref{fig:cloth-future-prediction} shows one interaction over its fitting and prediction frames.

\begin{table}[!htb]
    \centering
    \caption{\textbf{Future prediction.} All methods are fitted on the first 300 frames of eight red-cloth interactions and predict the rest of each recording under the recorded robot motion; each interaction is weighted equally.}
    \label{tab:future-prediction}
    \small
    \begin{tabular}{@{}lccccc@{}}
        \toprule
        Method & CD $\downarrow$ (mm$^2$) & IoU $\uparrow$ (\%) & PSNR $\uparrow$ (dB) & SSIM $\uparrow$ & LPIPS $\downarrow$ \\
        \midrule
        GS-Dynamics & 552.40 & 71.40 & 17.696 & 0.9132 & 0.1291 \\
        PhysTwin & 120.40 & 81.08 & 20.022 & 0.9316 & 0.0994 \\
        \method\ (ours) & \textbf{67.75} & \textbf{84.78} & \textbf{21.417} & \textbf{0.9396} & \textbf{0.0850} \\
        \bottomrule
    \end{tabular}
\end{table}

\subsection{Analysis}
\label{sec:analysis}

\begin{figure}[t]
    \centering
    \setlength{\tabcolsep}{1pt}
    \begin{tabular}{@{}cccc@{}}
        \includegraphics[width=0.235\linewidth]{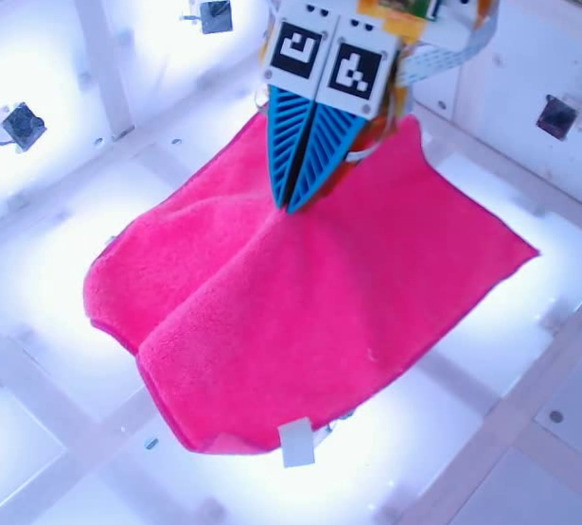} &
        \includegraphics[width=0.235\linewidth]{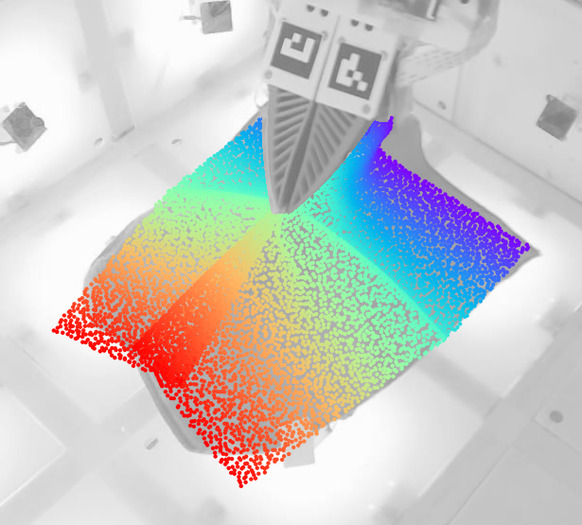} &
        \includegraphics[width=0.235\linewidth]{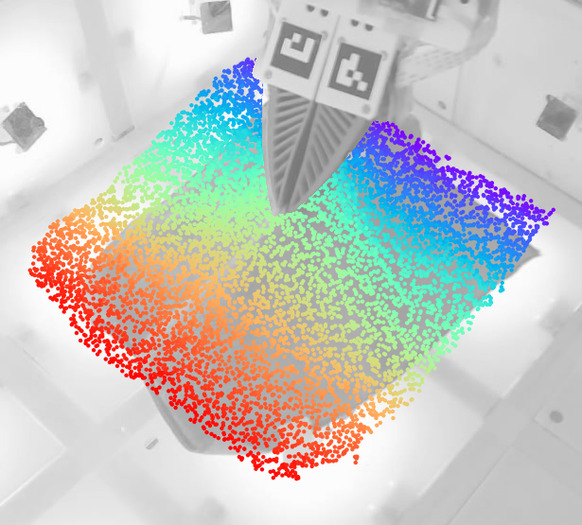} &
        \includegraphics[width=0.235\linewidth]{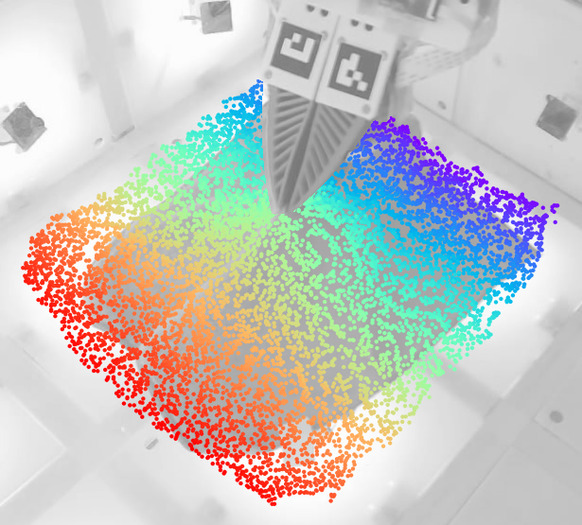} \\
        {\scriptsize (a) Observation} & {\scriptsize (b) \method} &
        {\scriptsize (c) PhysTwin, native springs} & {\scriptsize (d) PhysTwin + our grasp}
    \end{tabular}
    \caption{\textbf{Grasp transfer on one cloth interaction ($t=160$).} Our material-space grasp lifts the PhysTwin cloth near the fingertip (d) but does not recover the observed fold (a), which \method\ reproduces (b). Simulated panels show points sampled densely on the \method\ surface and the simulated PhysTwin particles, colored by their material coordinate at rest.}
    \label{fig:ms-grasp-ablation}
\end{figure}

\paragraph{Is the grasp model alone sufficient?}
Our comparisons change the physical representation and the grasp model together. To separate them, we transfer our grasp to PhysTwin: the gripper captures cloth particles, clamps them to gripper-local targets, and releases them, in place of the native controller springs, and this variant is fitted as in the main comparison (Appendix~\ref{app:ms-grasp-ablation}). The transferred grasp engages, but the sheet stays laterally spread and does not form the observed fold that \method\ reproduces (Figure~\ref{fig:ms-grasp-ablation}). The grasp model alone therefore does not account for the difference.

\paragraph{Can frictional contact replace the attachment?}
The gripper could instead hold the sheet through IPC frictional contact alone, without attachment. We explored this route and found two problems (Appendix~\ref{app:gradient}). First, the grasp is hard to form and slow to simulate. Grasps at an edge or a corner form readily, but grasps that must first pinch a fold out of the flat sheet slipped or stalled in our attempts unless the jaws followed a hand-tuned closing path. Even at an edge, whether the sheet stays in the jaws depends on the friction coefficient, and each forward pass is slow at the tolerance that accurate gradients require. Second, the gradients become unreliable once the lift begins: the adjoint agrees with finite differences while the fingers close, but not after the sheet is lifted, even with tightly converged solves. Our attachment instead holds the captured samples with a smooth energy from clamp to release, so retention does not depend on friction, and the grasp changes only at capture and release.

\subsection{Appearance-Aware Interactions}
\label{sec:applications}

The identified twin can be driven by actions that were never recorded. In Figure~\ref{fig:new-actions} (Appendix~\ref{app:new-actions}), the twin identified on one red-cloth interaction follows the recorded motion until the grasp closes; the gripper then lifts the cloth and either drapes it over a book or lowers it into an open box, and releases it. In a third scene, a free bottle stands on the cloth during a recorded two-gripper lift and topples as the cloth rises. The book, the box, and the bottle collide with the shell, and the Gaussian appearance follows the simulated shell without retraining. These scenes have no real counterpart, so the results are qualitative.

\section{Limitations}
\label{sec:limitations}

\paragraph{Identification and generalization.}
Following PhysTwin~\citep{jiang2025phystwin}, we identify each twin from a single recorded interaction, and the recovered parameters are effective values (Section~\ref{sec:method}). Our quantitative prediction therefore continues a recorded interaction, while the new actions in Figure~\ref{fig:new-actions} are evaluated only qualitatively. Identifying a shared object model from multiple interactions of the same object could better constrain its intrinsic properties and enable quantitative evaluation under unseen actions. In addition, the twin relies on tracked object points from calibrated multiview capture, and its accuracy is bounded by their quality; extending the pipeline to sparser observations with appropriate priors would make it applicable to less instrumented settings.

\paragraph{Appearance.}
Appearance is transported with the simulated shell rather than modeled jointly with it. The Gaussians therefore do not reproduce view-dependent shading on wrinkles, surfaces newly exposed by folding, or strain-dependent appearance, which limits the image metrics independently of the mechanics. Refining appearance jointly with the dynamics could improve the renderings, provided that the more expressive appearance model does not absorb errors of the physical model.

\section{Conclusion}

We presented \method, a physical digital twin for thin deformable objects that represents deformation and contact separately: a coarse bicubic Hermite shell models the object, and dense material-space grasp samples couple it to the robot. In experiments on real robot manipulation, \method\ resimulates observed motion and predicts held-out frames more accurately than spring--mass and learned baselines, and the identified twins can be driven by new actions and interact with new objects. Our analysis further suggests that the shell representation, not only the grasp model, contributes to this gain. Identifying a shared object model from multiple interactions and predicting its behavior under unseen actions are the next steps toward twins that support planning for contact-rich manipulation.

\section*{AI Use Statement}

Generative AI tools were used to assist with manuscript drafting, language editing, and checking clarity and consistency. The authors take full responsibility for the final content, including all claims, results, and references.

\bibliography{iclr2027_conference}
\bibliographystyle{iclr2027_conference}

\appendix

\section{Implementation Details}
\label{app:implementation}

\paragraph{Simulation.}
All twins use a $20\times20$ element grid, with nodal derivatives scaled by the element extent, and advance each video frame in five substeps. On the red cloth, grasp samples are spaced 2~mm apart at rest, giving about 21{,}000 candidates for a shell with 5{,}292 scalar degrees of freedom. The nominal material values are $E=100$~kPa, $\nu=0.3$, a density of 500~kg\,m$^{-3}$, and a thickness of $h=1$~mm for the red cloth and $h=5$~mm for the foam sheet; both stiffness scales in Eq.~(\ref{eq:elastic}) start at one. Linear systems are solved with a preconditioned iterative solver.

\paragraph{Optimization.}
Each system first evaluates its default parameters, then runs four CMA-ES generations of five candidates; failed candidates count toward this budget. The best feasible candidate initializes ten Adam iterations; for \method, each is one update of the five parameters, which share one Adam state. The selected model is the feasible forward run with the lowest fitting objective. In the main comparison, PhysTwin sums the spring forces on each particle in a fixed order, so repeated forward runs agree. All experiments run on NVIDIA RTX 5090 GPUs; fitting all ten interactions takes about 6.5~h for \method\ and 5~h for PhysTwin.

\section{Dataset}
\label{app:dataset}

We use ten interactions from Deform360~\citep{li2026deform360}: eight with a red cloth of about $29\times29$~cm and two with a white foam sheet of about $10.5\times15.7$~cm; the robot grasps the sheet with one or two grippers. Each interaction provides calibrated multiview video at 30 frames per second, the robot motion, and fingertip tactile signals. The tracked object points are obtained with the processing pipeline of Deform360. Contact onset, clamp, and release are detected by thresholding the tactile signals. Frame 0 initializes both systems and is not scored. The red-cloth interactions are fitted on frames 1--300, and the rest of each recording is held out for prediction; the foam interactions are fitted on frames 1--267.

\section{Grasp Transfer to a Mass--Spring Cloth}
\label{app:ms-grasp-ablation}

This study transfers the grasp model of Section~\ref{sec:grasp} to PhysTwin on a red-cloth interaction. Both PhysTwin variants keep the spring--mass model, integrator, and optimizer of the main comparison, and both are fitted with the objective of Eq.~(\ref{eq:geomloss}) and the optimization budget used for Tables~\ref{tab:reconstruction-resimulation} and~\ref{tab:future-prediction}. The native variant drives the cloth through controller springs, as in the main comparison. The transferred-grasp variant replaces them with our grasp model, applied to the particles in place of the material samples of \method: every cloth particle is a grasp candidate, captured when the thin box around a gripper pad sweeps over it; its gripper-local target relaxes before clamp, is frozen at clamp, and is cleared on release; and the total grasp stiffness is divided by the number of active particles, as in Eq.~(\ref{eq:sample_attach_energy}). Capture, clamp, and release therefore follow the same pad geometry and the same recorded contact events as in \method, and the variant additionally optimizes its grasp stiffness.

The captured particles follow the gripper, but the sheet stays laterally spread instead of forming the observed fold that \method\ reproduces (Figure~\ref{fig:ms-grasp-ablation}).

\section{Grasping through Frictional Contact}
\label{app:gradient}

We also tried to hold the sheet through contact alone, without attachment, using either penalty contact forces or IPC~\citep{li2020ipc} with friction between the shell and the fingers, which follow the recorded gripper motion. Both variants share two problems. First, the grasp often fails to lift the sheet. Grasps at an edge or a corner can enter the jaws directly, but grasps that must first pinch a fold out of the sheet are fragile: for one gripper of a bimanual grasp, the sheet slips or the Newton solve stalls across variations of jaw opening, pose, height, and bending stiffness, unless the jaws are pre-opened and closed along a staged, hand-tuned path that lets the sheet buckle into a load-bearing fold. Grasping through collision also takes at least three times as long to simulate as our attachment. Second, the gradients are inaccurate: the adjoint agrees with finite differences while the fingers close, but not after the sheet is lifted. We suspect two causes. Coupling the shell to the fingers as rigid bodies~\citep{lan2022abd} degrades the linear systems that the adjoint replays, and when the shell is lifted through friction alone, the contact locations change too abruptly between steps for the rollout to stay smooth in the parameters. Our attachment avoids both problems, since it holds the captured samples with a smooth energy that changes only at capture and release.

\section{Driving the Twin with New Actions and Objects}
\label{app:new-actions}

\begin{figure}[!t]
    \centering
    \setlength{\tabcolsep}{1pt}
    \renewcommand{\arraystretch}{0.5}
    \begin{tabular}{@{}cccccc@{}}
        \includegraphics[width=0.162\linewidth]{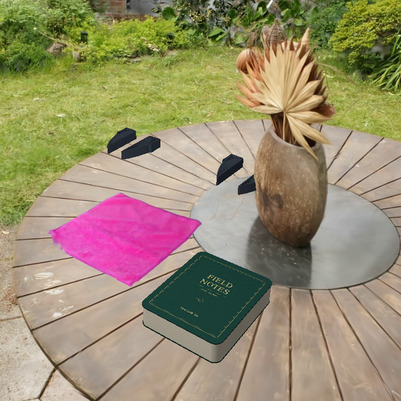} &
        \includegraphics[width=0.162\linewidth]{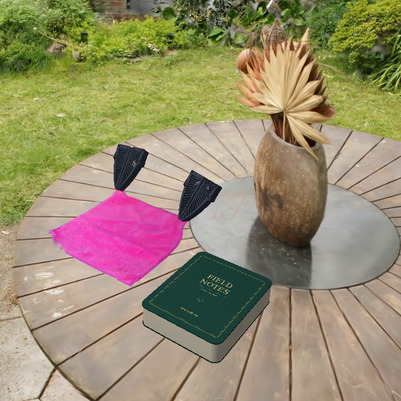} &
        \includegraphics[width=0.162\linewidth]{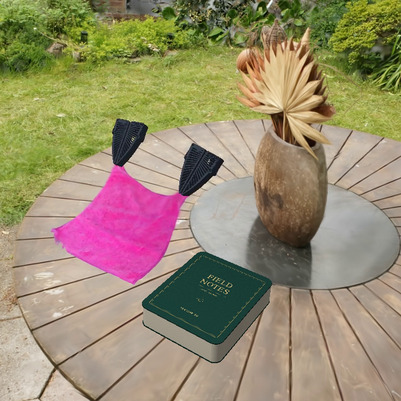} &
        \includegraphics[width=0.162\linewidth]{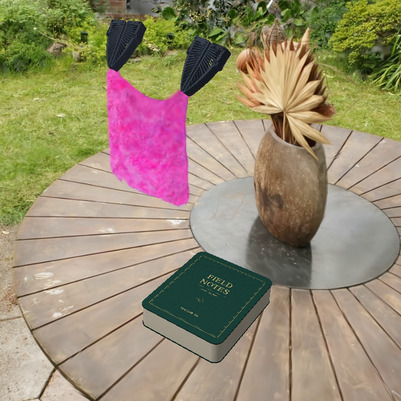} &
        \includegraphics[width=0.162\linewidth]{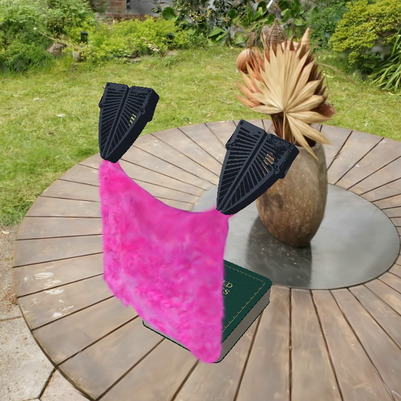} &
        \includegraphics[width=0.162\linewidth]{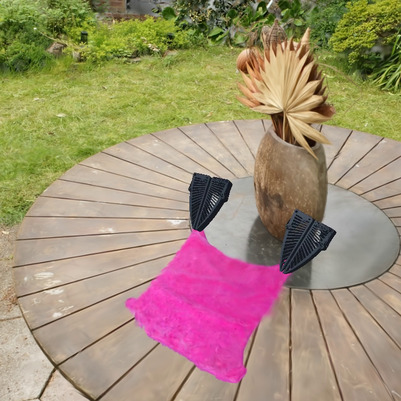} \\
        \includegraphics[width=0.162\linewidth]{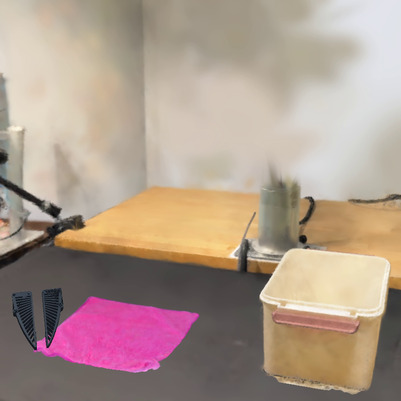} &
        \includegraphics[width=0.162\linewidth]{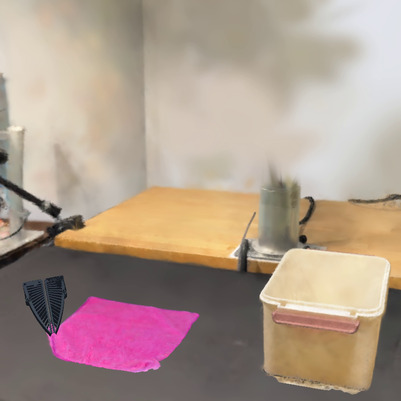} &
        \includegraphics[width=0.162\linewidth]{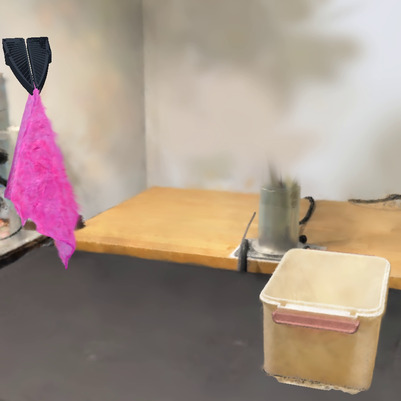} &
        \includegraphics[width=0.162\linewidth]{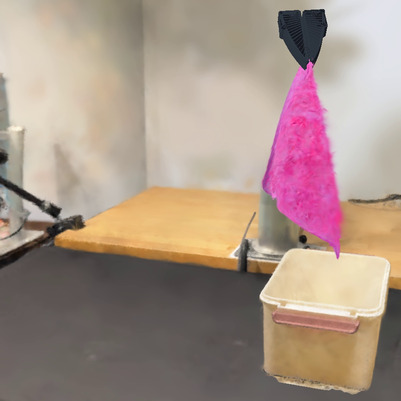} &
        \includegraphics[width=0.162\linewidth]{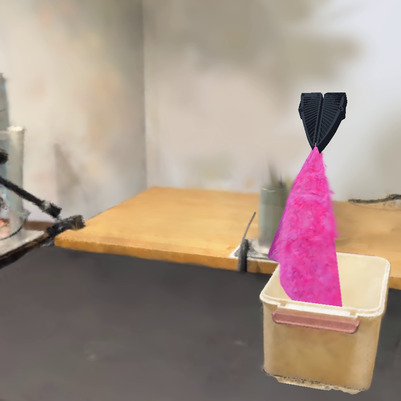} &
        \includegraphics[width=0.162\linewidth]{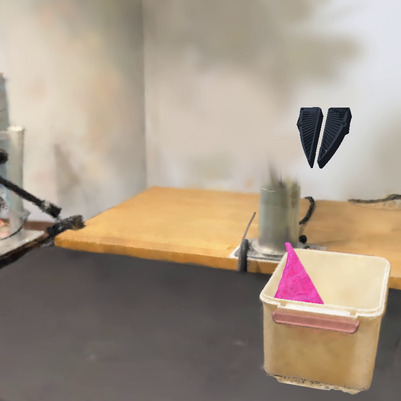} \\
        \includegraphics[width=0.162\linewidth]{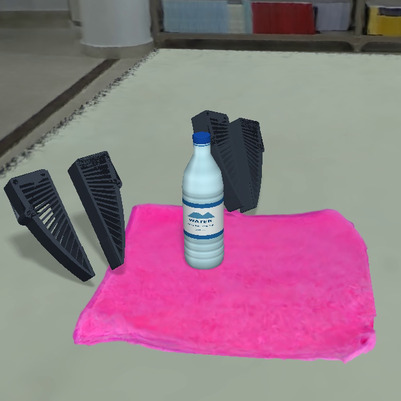} &
        \includegraphics[width=0.162\linewidth]{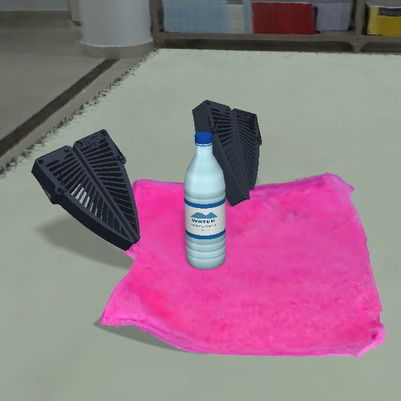} &
        \includegraphics[width=0.162\linewidth]{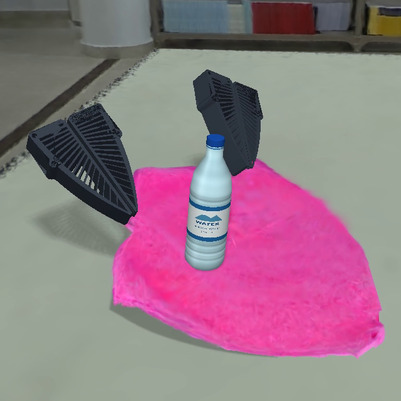} &
        \includegraphics[width=0.162\linewidth]{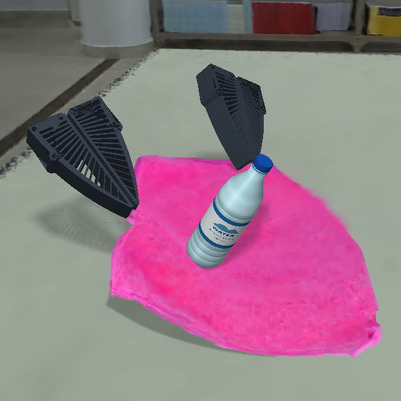} &
        \includegraphics[width=0.162\linewidth]{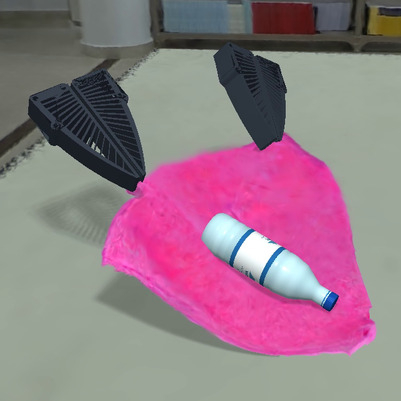} &
        \includegraphics[width=0.162\linewidth]{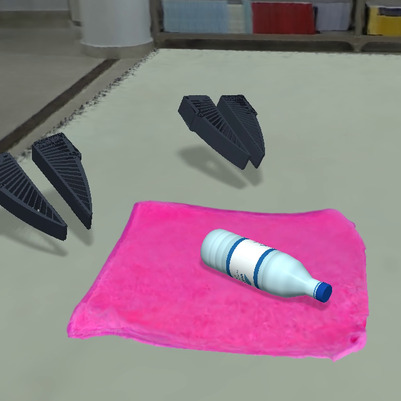}
    \end{tabular}
    \caption{\textbf{Driving the identified cloth with new actions and objects.} Top: draping the cloth over a book. Middle: placing the cloth inside a box. Bottom: a free bottle toppling as a recorded two-gripper lift raises the cloth.}
    \label{fig:new-actions}
\end{figure}

All scenes use twins identified on red-cloth interactions, with unchanged parameters (Figure~\ref{fig:new-actions}). In the first two, the recorded gripper motion is kept until the grasp closes and then replaced by new motions that drape the cloth over a book or lower it into an open box; in the third, a recorded two-gripper lift is replayed with a free bottle standing on the cloth. The book and the box are fixed obstacles and the bottle is an affine body~\citep{lan2022abd}, all colliding with the shell through IPC. The cloth is rendered with its reconstructed Gaussians transported by the simulated shell.

\makeatletter\setlength{\@fptop}{0pt}\makeatother

\end{document}